\documentclass{article}

\usepackage[english]{babel}

\usepackage[letterpaper,top=2cm,bottom=2cm,left=3cm,right=3cm,marginparwidth=1.75cm]{geometry}

\usepackage{amsmath}
\usepackage{booktabs} % 必须导入这个包来画三线表
\usepackage{multirow} % 如果有合并行需要这个包
\usepackage{graphicx}
\usepackage{subcaption}
\usepackage[colorlinks=true, allcolors=blue]{hyperref}
\usepackage{listings}

\title{Adaptive Dueling Double Deep Q-networks in Uniswap V3\\
Replication and Extension with Mamba}
\date{}
\author{Zhaofeng Zhang\\
Department of Mathematics, University of Michigan}

\begin{document}
\maketitle
\section*{Summary}
The report goes through the main steps of replicating and extending the article “Adaptive Liquidity Provision in Uniswap V3 with Deep Reinforcement Learning.” The replication part includes how to obtain data from the Uniswap Subgraph, details of the implementation, and comments on the results. After the replication, we propose a new structure based on the original model, which combines Mamba with DDQN and a new reward function. In this new structure, we clean the data again and introduce two new baseline methods for comparison. As a result, although the model has not yet been applied to all datasets, it shows stronger theoretical support than the original model and performs better in the baseline tests.\\
\textbf{Keywords:} Reinforcement Learning, Double Deep Q-Learning, Uniswap V3, Mamba, Market Maker

\section*{Introduction}
%\subsection{General Goal of the paper}
The original work \cite{zhang2023adaptive} proposed a structure based on Dueling DDQN \cite{wang2016dueling} on the record of ETH/USDC-$0.3\%$ and ETH/USDT-$0.3\%$ pool on the Uniswap V3  \cite{adams2021uniswapv3}. The \textbf{goal} of the work is to design a strategy based on the enhanced Dueling DDQN to control the liquidity risk optimally.The model constructed in the paper considers 28 features extracted from the Uniswap Subgraph and introduces new components for the state representation, action–value streams, and reward function. The training, validation, and testing datasets span the period from 2021/08/02 to 2023/01/25 (divided into four sections), and the model was compared against three baseline methods. Based on the reported results, the proposed model outperformed all baselines in terms of gas cost, trading fees, Loss-Versus-Rebalancing (LVR), and Profit-and-Loss (PnL).

\begin{figure}[b]
    \centering
    \includegraphics[width=0.8\linewidth]{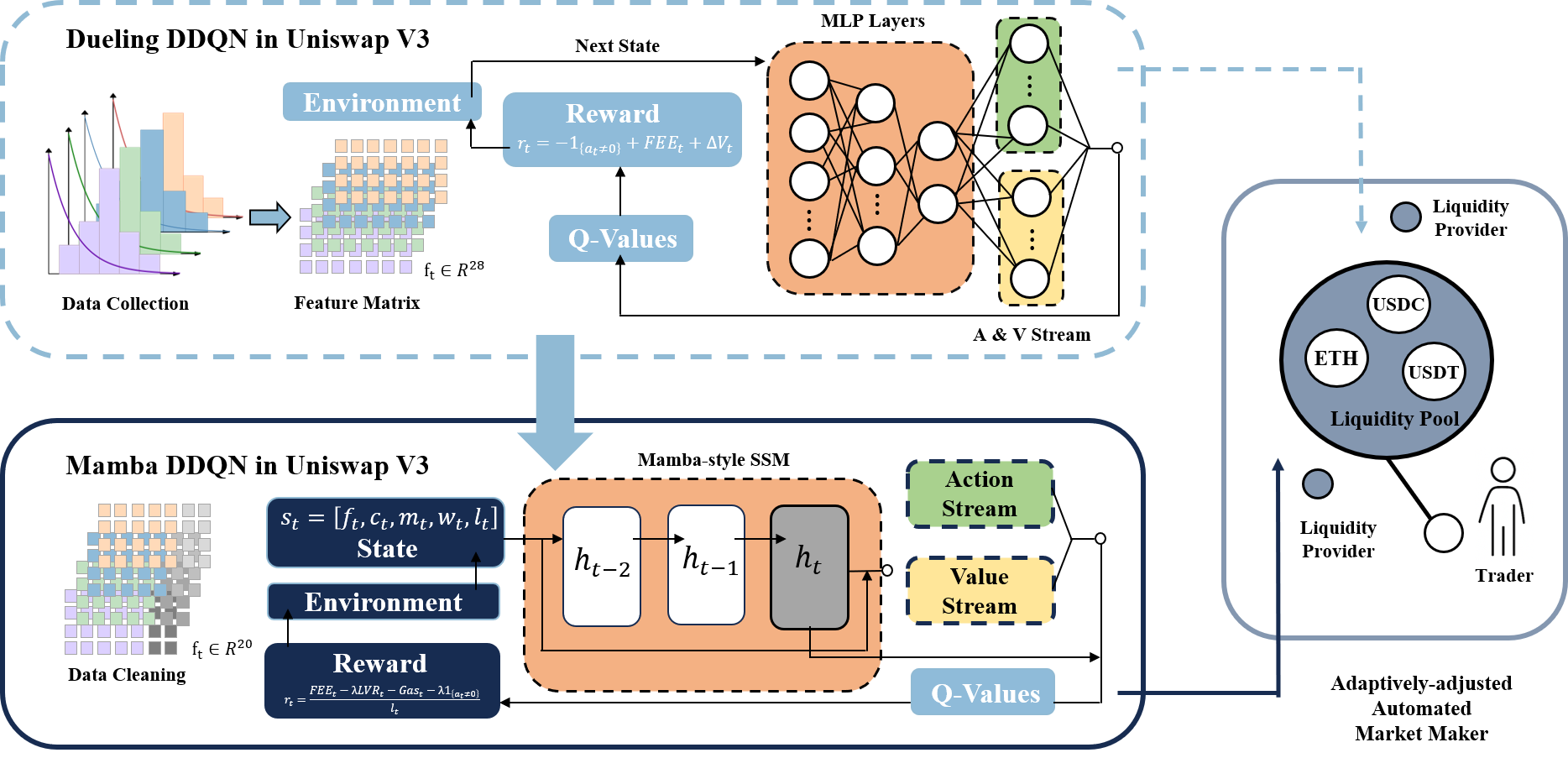}
    \caption{Overview of the work}
    \label{Overview}
\end{figure}
 
The purpose of this report is to replicate the original work and explore possible extensions. The overall workflow is illustrated in Figure~\ref{Overview}. The structure of the report is organized as follows. Section~\ref{PaperReview} reviews the replication process, including data sources, implementation details, results, and discussion. Section~\ref{Extension} presents a proposed extension of the original model, named Mamba-DDQN, along with the methodology, comparisons, and discussion. A broader reflection on the entire study is provided in Section~\ref{Discussion}.

\section{Paper Review: Dueling DDQN in Uniswap V3}
\label{PaperReview}
%EMH to market roles
\subsection{Data Sources Replication}
\label{process}
To replicate the work, historical data of the pool ETH/USDC-$0.3\%$ and ETH/USDT-$0.3\%$ from 2021 to 2023 will be collected and used. The dataset used for replication was collected from \textit{The Graph}\footnote{https://thegraph.com/explorer/subgraphs/5zvR82QoaXYFyDEKLZ9t6v9adgnptxYpKpSbxtgVENFV} as the link provided by the author is unavailable now. The Graph is also an open-source, decentralized protocol that records the indexing and querying of blockchain data.

\begin{figure}[t]
    \centering
    \begin{subfigure}{0.45\textwidth}
        \centering
        \includegraphics[width=\linewidth]{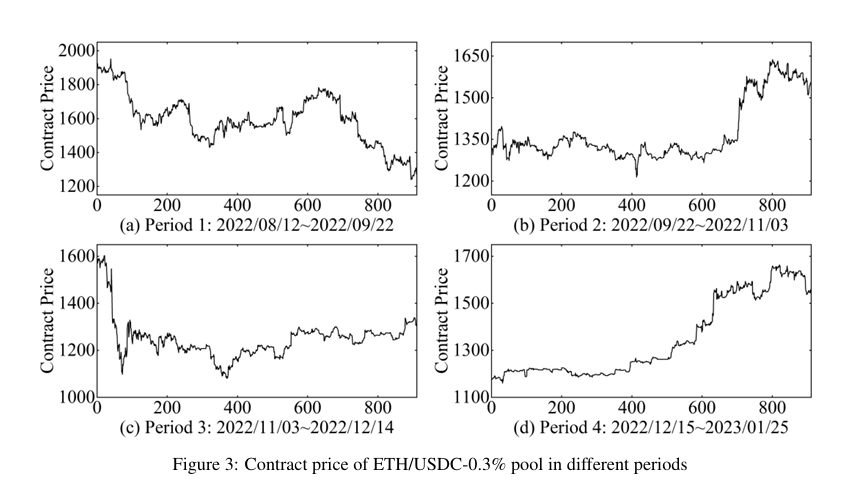}
        \caption{Origin figure}
        \label{fig:1}
    \end{subfigure}
    \hfill
    \begin{subfigure}{0.38\textwidth}
        \centering
        \includegraphics[width=\linewidth]{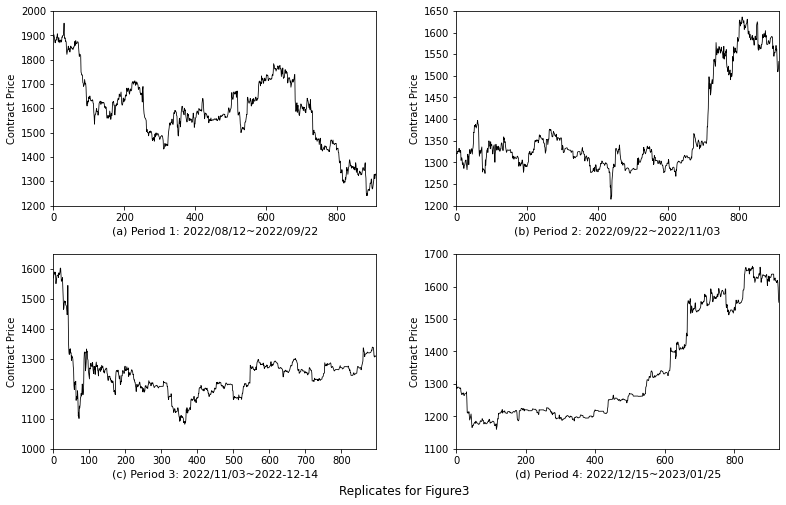}
        \caption{Replicated }
        \label{fig:2}
    \end{subfigure}
    \caption{Comparison between the s of contract price}
    \label{price}
\end{figure}

However, the original Uniswap V3 data only includes some basic information, such as $"tick"$, $"liquidity"$ and $"sqrtPrice"$. So we need to calculate some features mentioned in the paper, such as $contract\_price$, $"cci\_14"$, $smi\_1$ and $"stoch_d"$. Unfortunately, The Graph seems to have failed to provide one-hand volume data at the hourly tick-level, so we recalculated and added them as a new column. Figure \ref{price} shows the original and replicated figure for the contract price shown in the paper ("Figure 3" in the original paper), which is almost the same.

The next step is data cleaning and then dividing the set for training and testing. In the data cleaning, each time tick only contains once, and more raw data will be removed. The rows containing a NaN value were also removed. After the data cleaning, there are 25,649 rows in the dataset (before cutting to set). And the dataset will be divided into four periods and each period includes nearly 13 months of data. Table \ref{set} shows the hours included in each set, which is very close to the requirement in the paper (8000 for training, 1000 for validation, 1000 for testing).

\begin{table*}[ht]
\centering
\small
\begin{tabular}{lcccccccc}
\toprule
 & \multicolumn{3}{c}{\textbf{ETH/USDC-0.3\%}} & \multicolumn{3}{c}{\textbf{ETH/USDT-0.3\%}} \\
\cmidrule(lr){2-4} \cmidrule(lr){5-7}
 & Train & Valid & Test & Train & Valid & Test \\
\midrule

Period 1 2021/08/02- 2022/09/22 & 7983 & 984 & 984 & 7964 & 984 & 984 \\
Period 2 2021/09/12- 2022/11/03 & 7983 & 984 & 1008 & 7972 & 984 & 983 \\
Period 3 2021/10/24- 2022/12/14 & 7983 & 1008 & 984 & 7973 & 984 & 976\\
Period 4 2021/12/05- 2023/01/25 & 7984 & 984 & 981 & 7958 & 984 & 954 \\
\midrule
Paper Provided & 8000 & 1000 & 1000 & 8000 & 1000 & 1000\\

\bottomrule
\end{tabular}
\caption{Hours for the train, valid, test set.}
\label{set}
\end{table*}

\subsection{Implementation Detail}
\label{Im}
%state variables, action space, reward function, and training procedure)
Basically speaking, this paper is working on the previous work of Dueling DDQN \cite{wang2016dueling}, but making some changes in the state, action, and reward to better suit the features of the market. 

\subsubsection{State}
The model defined a state $s_t$ at time $t$:
$$
s_t = [\textbf{f}_t, c_t, m_t, w_t, l_t] \in \mathbf{R}^{32}, \textbf{f}_t \in \mathbf{R}^{28}
$$
where at time $t$, $\textbf{f}_t$ is the feature collected in Section \ref{process}, $c_t$ is the amount of USD the model holds, $m_t$ is the tick of the liquidity position, $w_t$ is the width of the liquidity interval and $l_t$ is the value of the fund. In the code, it is:
\begin{lstlisting}[basicstyle=\ttfamily\small]
state = np.concatenate([f_vec, [c_t, m_t, w_t, l_t]])
\end{lstlisting}

\subsubsection{Action Space}
The action space is defined as a set:
$$
A = \{0, 1, \ldots, N_a\}
$$
where $N_a$ is the maximum width of the interval. When $a =0$, the agent will not do anything and the agent will create a new liquidity range with the width $w_{t+1} = a_t$. In the implemented code:
\begin{lstlisting}[basicstyle=\ttfamily\small]
if action == 0:
    m_t1 = self.m_t
    w_t1 = self.w_t
else:
    tick_now = price_to_tick(p_t)
    m_t1 = d * round(tick_now / d)
    w_t1 = float(action)
\end{lstlisting}

\subsubsection{Reward Function}
The paper defined a new reward function that considers the LVR and gas fee, which is:
$$
r_t = -\textbf{1}_{\{a_t \neq 0\}} + Fee_t + \Delta LVR_t
$$
where at time t, $r_t$ is the obtained reward and $\textbf{1}_{\{a_t \neq 0\}}$ reflects whether the gas fees had incurred. $Fee_t$ and $\Delta LVR_t$ are the gas fee and LVR computed.

\subsubsection{Trading Fee and Gas Fee}
Based on the rules of Uniswap V3, the trading fee can be represented in this way:
$$
fee = \{ 
\begin{array}{cc}
     \frac{\delta}{1-\delta}L (\sqrt{p'} - \sqrt{p}) & p \leq p'  \\
    \frac{\delta}{1-\delta}L (\sqrt{p'} - \sqrt{p}) & p \geq p'
\end{array}
$$

\noindent And the gas fee is the flat fee per reallocation. In the code, we have:
\begin{lstlisting}[basicstyle=\ttfamily\small]
fee_pool = row_t1[FEE_COL]
fee_agent = fee_pool * share_t

#gas fee
gas_t = self.gas_flat if action != 0 else 0
\end{lstlisting}

\subsubsection{Update and Transition}
The change of the value for each state is shown below, where $r_t$ is the gained reward and $l_t$ is the current fund value at time $t$:
$$
l_{t+1} = l_t + r_t
$$

For the transition part, there are several steps during the transition. First, convert the market price $c_t$ to a tick:
$$
m_{t+1} \leftarrow d \cdot round(price2tick(c_t), d)
$$
Second, the width will be renewed as $w_{t+1} \leftarrow a_t$, where $a_t$ decides the width of the next range. And the next step is computing the liquidity units $u$ by $c_t + l_t$, $m_{t+1}$, $w_{t+1}$ and $close_t$, which means the LP get $c_t$ and quit from the old range. Then use $c_t$ to get the new LP position, that is, get the liquidity units $u$.

After getting the new liquidity unit, the agent will go through the environment and move forward to the next time $\textbf{f}_{t+1}$, renew $c_{t+1} = c_t + Fee_t$ and renew the value of LP $l_{t+1}$. 

\subsection{Modeling Approach}
\label{approach}
\begin{figure}
    \centering
    \includegraphics[width=\linewidth]{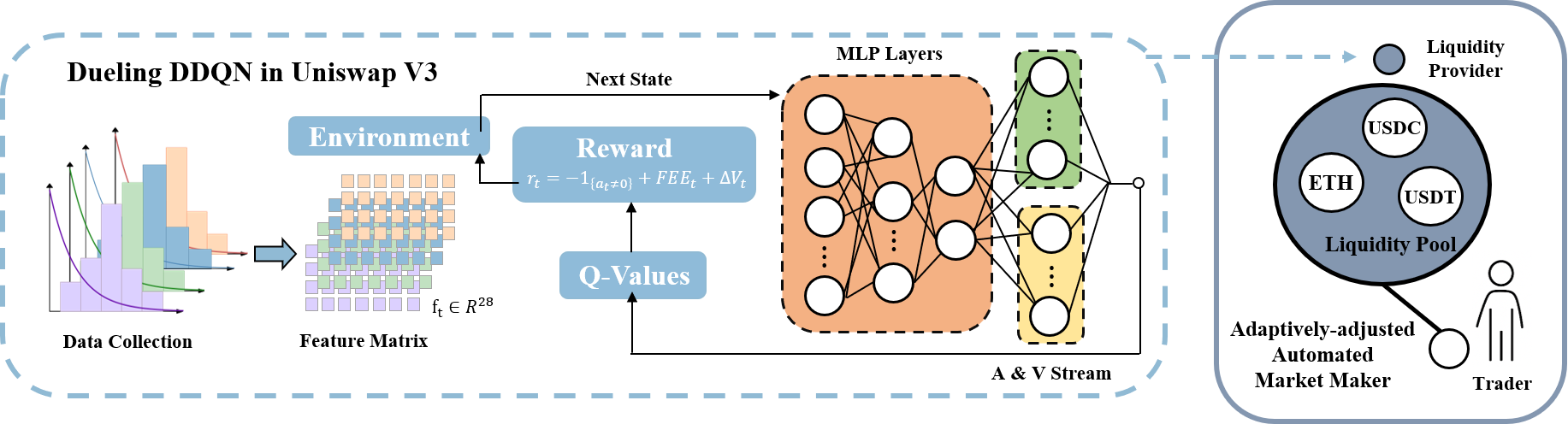}
    \caption{Caption}
    \label{Framework of DDQN}
\end{figure}
\subsubsection{Experience Replay}
The experience replay module is based on the work of \cite{mnih2013playing}. During the training process, the transitions will be stored in a replay buffer $D$. When renewing the Deep Q-learning Network (DQN), there will be a one-time random sampling minibatch:
$$
\{(s_i, a_i, r_i, s_i^{'}, d_i)\}^{B}_{i=1} \sim Uniform (D) 
$$
We implemented this in the following code:
\begin{lstlisting}[basicstyle=\ttfamily\small]
replay_buffer.push(state, action, reward_rel, next_state, float(done))
batch = replay_buffer.sample(BATCH_SIZE)
\end{lstlisting}

\subsubsection{Dueling DDQN}
The framework of the Dueling DDQN is shown in Figure \ref{Framework of DDQN}. In the Section \ref{Im}, we go through the implementation details of the Dueling DDQN model. For a value function $V(s)$ and an advantage function $A(s,a)$, we have:
$$
Q(s,a) = V(s) + (A(s,a) - \frac{1}{|A|}\sum_{a'}A(s,a'))
$$
Then there is the policy network for the action and the target network for the value:
$$
a^* = arg\max_a Q_{policy}(s',a), \quad y = r + \gamma Q_{target}(s', a^*)
$$
In the code, we have:
\begin{lstlisting}[basicstyle=\ttfamily\small]
value = self.value_stream(x)
adv = self.adv_stream(x)
q = value + (adv - adv.mean(dim=1))
\end{lstlisting}

\noindent The hyperparameters of the Dueling DDQN model are given in the Table \ref{DDQN Parameter}.

\begin{table}[]
    \centering
    \begin{tabular}{c|c}
    \hline
    Hyperparameters & Values \\
    \hline
 Hidden Unites & [64,64] \\
 Activation & ReLU\\
 Final Activation & None\\
 Learning Rate & $10^{-4}$\\
 Batch Size & 256\\
 Buffer Size & $10^6$\\
 Discounted Factor& 0.9\\
 Target Update Rate & 0.01\\
 Gradient Clipping Norm & 0.7\\
    \hline
    \end{tabular}
    \caption{Hyperparameters of Dueling DDQN}
    \label{DDQN Parameter}
\end{table}

\subsubsection{Baseline M2: Uniform $\tau$-rest Strategy}
M2 is based on the work of \cite{fan2021strategic}. In the uniform $\tau$-rest strategy, the width of the range $\tau$ is a fixed value and LP will provide a centered at the current price $[m_t-d_{\tau}, m_t + d_{\tau}]$. This strategy only acts when the price runs out of the range and does not consider the gas fee very much. The value of $\tau$ is provided in the paper, and it is shown in the Table \ref{value of tau}.

\begin{table}[htbp]
\centering
\caption{Hyperparameters of Uniform $\tau$-reset Strategy}
\begin{tabular}{l c c c c}
\hline
$\tau$ & & $l_0 = 250$ & $l_0 = 500$ & $l_0 = 1000$ \\
\hline

\multirow{4}{*}{\begin{tabular}[c]{@{}l@{}}ETH-USDC \\ 0.3\%\end{tabular}}
& Period 1 & 6 & 4 & 1 \\
& Period 2 & 5 & 2 & 1 \\
& Period 3 & 6 & 3 & 2 \\
& Period 4 & 4 & 3 & 1 \\
\hline

\multirow{4}{*}{\begin{tabular}[c]{@{}l@{}}ETH-USDT \\ 0.3\%\end{tabular}}
& Period 1 & 6 & 4 & 1 \\
& Period 2 & 5 & 2 & 1 \\
& Period 3 & 10 & 3 & 1 \\
& Period 4 & 4 & 3 & 1 \\
\hline

\end{tabular}
\label{value of tau}
\end{table}

\subsubsection{Baseline M3: Exponential Weights Adaptive Strategy (EWA)}
M3 is the method of Exponential Weights Adaptive Strategy (EWA)\cite{bar2023uniswap}, which is a bandit method to make an action in a potential range with probability. Give $N$ numbers of potential range $n \in \{1, \ldots, N\}$, the renew of the weight is:
$$
p_t(n) = \frac{exp(\eta\sum^t_{\tau=1} r_{\tau}(n))}{\sum^N_{\mu=1}exp(\eta\sum^t_{\tau=1} r_{\tau}(\mu))}
$$
where $\eta$ is the learning rate, $r_t(n)$ is the return based on the width $n$ at time $t$. The parameters are shown in the Table \ref{parameter of EWA}.

\begin{table}[htbp]
\centering
\caption{Hyperparameters of EWA}

\begin{tabular}{l c c c c c c c c c c }
\hline
 & & \multicolumn{3}{c}{$l_0 = 250$} & \multicolumn{3}{c}{$l_0 = 500$} & \multicolumn{3}{c}{$l_0 = 1000$} \\
 & & $N$ & $\eta$ & $T_{re}$ & $N$ & $\eta$ & $T_{re}$ & $N$ & $\eta$ & $T_{re}$ \\
\hline

\multirow{4}{*}{ETH-USDC 0.3\%}
& Period 1 & 10 & 1 & 21 & 10 & 1 & 14 & 10 & 1 & 6 \\
& Period 2 & 10 & 10 & 24 & 10 & 10 & 24 & 10 & 10 & 9 \\
& Period 3 & 10 & 1 & 22 & 10 & 4 & 15 & 10 & 1 & 13 \\
& Period 4 & 10 & 7 & 24 & 10 & 1 & 21 & 10 & 1 & 18 \\
\hline

\multirow{4}{*}{ETH-USDC 0.3\%}
& Period 1 & 10 & 1 & 21 & 10 & 1 & 14 & 10 & 1 & 6 \\
& Period 2 & 10 & 10 & 24 & 10 & 10 & 24 & 10 & 10 & 12 \\
& Period 3 & 10 & 1 & 22 & 10 & 7 & 22 & 10 & 10 & 3 \\
& Period 4 & 10 & 7 & 21 & 10 & 1 & 21 & 10 & 1 & 21 \\
\hline

\end{tabular}
\label{parameter of EWA}
\end{table}

\subsubsection{Baseline M4: Dynamic Programming}
M4 is the method that is based on predictable loss and optimal LP provision \cite{cartea2024decentralized}. In the dynamic programming model, LVR can be represented as :
$$
LVR = \int \sigma^2(\cdot)dt
$$
So we can consider it as an optimal control problem, which the objective is shown below:
$$
\max_{p_t \in range} \mathbf{E}[Fee_t - LVR_t]
$$
And we can find the optimal solution subject to:
$$
V(p) = \max_{(l,r)}\{Fee(p,l,r) + \mathbf{E}[V(p_{t+1})]\}
$$

\section{Replication Result}
\subsection{Result}
Based on the code and methods shown in Section \ref{Im} and \ref{approach}, we tried to replicate the result of the paper. Tables \ref{performance1} and \ref{performance2} show the performance between the original result and replicated result for the ETH/USDC-$0.3\%$ and ETH/USDC-$0.3\%$ pools, respectively. 

Although they are not perfectly fixed to the original result, the replicated result still shows some same features. Both the PnL of M1 perform best of all in all the periods and the LVR of M1 is always the highest in all the periods. The gas fee is the best of M1, M2, and M3 for most cases. These trends appear in both the pool of ETH/USDC-$0.3\%$ and ETH/USDT-$0.3\%$.

However, there are some unsatisfied with some replicated results. The result of the relative gas fee of M4 is too small to influence the final comparison. M3 seems to perform vary from the original result in relative trading fee.

\begin{table*}[ht]
\centering
\small
\begin{tabular}{llcccc|cccc}
\toprule
& \textbf{ETH/USDC-0.3\%} & \multicolumn{4}{c}{\textbf{Original Result}} & \multicolumn{4}{c}{\textbf{Replicated Result}} \\
\cmidrule(lr){3-6} \cmidrule(lr){7-10}
& & M1 & M2 & M3 & M4 & M1 & M2 & M3 & M4 \\
\midrule

\textbf{Period 1} & relative trading fee $\uparrow$ & 0.691 & 0.630 & \textbf{0.774} & 0.088 & 0.501 & \textbf{0.873} & 0.455 & 0.464  \\
& relative gas fee $\downarrow$    & \textbf{0.113} & 0.207 & 0.293 & 0.120 &  0.155 & 0.337 & 0.184 & \textbf{0.046} \\
& relative LVR $\downarrow$        & 0.205 & 0.205 & 0.248 & \textbf{0.030} & 0.339  & 0.165 & 0.094 & \textbf{0.060} \\
& relative PnL $\uparrow$        & \textbf{0.373} & 0.218 & 0.232 & -0.210 & \textbf{0.421}  & 0.218 & 0.177 & -0.06 \\

\midrule
\textbf{Period 2} & relative trading fee $\uparrow$ & 0.607 & 0.453 & \textbf{0.611} & -0.052 & 0.752  & \textbf{0.872} & 0.392 & 0.009 \\
& relative gas fee $\downarrow$    & \textbf{0.118} & 0.153 & 0.253 & 0.120 &  0.192 & 0.303 & 0.164 & \textbf{0.003} \\
& relative LVR $\downarrow$        & 0.182 & 0.132 & 0.168 & \textbf{0.015} & 0.217  & 0.129 & 0.065 & \textbf{0.035} \\
& relative PnL $\uparrow$        & \textbf{0.307} & 0.168 & 0.190 & -0.083 & \textbf{0.250}  & 0.185 & 0.163 & -0.029 \\

\midrule
\textbf{Period 3} & relative trading fee $\uparrow$& 0.541 & \textbf{0.569} & 0.550 & 0.089 & 0.530 & \textbf{0.666} & 0.283 & 0.020  \\
 & relative gas fee $\downarrow$    & \textbf{0.104} & 0.167 & 0.307 & 0.124 & 0.156 & 0.289 & 0.176 &  \textbf{0.010} \\
 & relative LVR $\downarrow$       & 0.220 & 0.257 & 0.174 & \textbf{0.045} & 0.137 & 0.135 & 0.085 & \textbf{0.096} \\
 & relative PnL $\uparrow$        & \textbf{0.217} & 0.145 & 0.096 & -0.079 & \textbf{0.375} & 0.172 & 0.022 &  -0.087 \\

\midrule
\textbf{Period 4} & relative trading fee $\uparrow$& 0.370 & 0.318 & \textbf{0.365} & 0.031 & 0.254  & \textbf{0.319} & 0.279 &  0.020\\
 & relative gas fee $\downarrow$    & \textbf{0.064} & 0.100 & 0.253 & 0.120 & 0.134  & 0.212 & 0.156 & \textbf{0.037} \\
 & relative LVR $\downarrow$        & 0.121 & 0.108 & 0.107 & \textbf{0.014} &  0.162 &  0.095 & 0.044 & \textbf{0.020} \\
 & relative PnL $\uparrow$        & \textbf{0.185} & 0.109 & 0.004 & -0.104 & \textbf{0.320}  & 0.156 & 0.079 &  -0.04 \\

\bottomrule
\end{tabular}
\caption{Replicated result for ETH/USDC-$0.3\%$}
\label{performance1}
\end{table*}

\begin{table*}[ht]
\centering
\small
\begin{tabular}{llcccc|cccc}
\toprule
& \textbf{ETH/USDT-0.3\%} & \multicolumn{4}{c}{\textbf{Original Result}} & \multicolumn{4}{c}{\textbf{Replicated Result}} \\
\cmidrule(lr){3-6} \cmidrule(lr){7-10}
& & M1 & M2 & M3 & M4 & M1 & M2 & M3 & M4 \\
\midrule

\textbf{Period 1} & relative trading fee $\uparrow$&  0.645 & 0.614 & \textbf{0.921} & 0.402 & 0.567 & \textbf{0.779} & 0.450 &  0.041 \\
& relative gas fee $\downarrow$    & \textbf{0.096} & 0.207 & 0.293 & 0.133 & 0.157  & 0.336 & 0.184 & \textbf{0.045} \\
& relative LVR  $\downarrow$     & 0.199 & 0.211 & 0.316 & \textbf{0.139} & 0.288  & 0.148 & 0.092 & \textbf{0.052} \\
& relative PnL $\uparrow$         & \textbf{0.310} & 0.196 & 0.131 & 0.130 & \textbf{0.363}  & 0.192 & 0.174 & -0.056 \\

\midrule
\textbf{Period 2} & relative trading fee $\uparrow$ & 0.503 & 0.459 & \textbf{0.587} & 0.452 & 0.752  & \textbf{0.964} & 0.421 &  0.009\\
& relative gas fee $\downarrow$     & \textbf{0.089} & 0.147 & 0.253 & 0.140 & 0.128  & 0.307 & 0.164 & \textbf{0.030} \\
& relative LVR  $\downarrow$       & 0.161 & 0.159 & 0.186 & \textbf{0.125} & 0.210  & 0.143 & 0.072 & \textbf{0.037} \\
& relative PnL $\uparrow$         & \textbf{0.252} & 0.154 & 0.178 & 0.187 & 0.126  & \textbf{0.208} & 0.185 & -0.031 \\

\midrule
\textbf{Period 3} & relative trading fee $\uparrow$ & 0.686 & 0.707 & \textbf{0.774} & 0.295 & \textbf{0.811}  & 0.564 & 0.265 & 0.197 \\
 & relative gas fee $\downarrow$     & \textbf{0.119} & 0.060 & 0.207 & 0.094 & 0.173  & 0.263 & 0.172 & \textbf{0.120} \\
 & relative LVR  $\downarrow$        & 0.335 & 0.209 & 0.314 & \textbf{0.103} &  0.142 & 0.131 & \textbf{0.088} &  0.092 \\
 & relative PnL $\uparrow$         & \textbf{0.232} & 0.122 & 0.109 & 0.097 &  \textbf{0.399} & 0.143 & 0.004 & -0.079  \\

\midrule
\textbf{Period 4} & relative trading fee $\uparrow$ & 0.463 & 0.348 & \textbf{0.373} & 0.149 & 0.259 & \textbf{0.350} & 0.271 &  0.005 \\
 & relative gas fee $\downarrow$     & \textbf{0.066} & 0.107 & 0.293 & 0.120 & 0.148  & 0.214 & 0.176 &  \textbf{0.060} \\
 & relative LVR $\downarrow$         & 0.271 & 0.159 & 0.115 & \textbf{0.059} &  0.230 &  0.101 & 0.043 &  \textbf{0.010}\\
 & relative PnL $\uparrow$         & \textbf{0.124} & 0.082 & -0.035 & -0.030 & \textbf{0.393}  & 0.164 & 0.052 &  -0.011 \\

\bottomrule
\end{tabular}
\caption{Replicated result for ETH/USDT-$0.3\%$}
\label{performance2}
\end{table*}

\subsection{Comments for Result}
\label{comment}
Based on the difference between the original result and the replicated result, here are several potential reasons.
\subsubsection{Data Differences}
The record data of Uniswap may be different from the data screenshots taken by the authors, or the data we collected may still not the 100\% similar to the data that the authors had used. Figure \ref{price} looks very same, but there are slight differences between the original one and mine. One possible reason is that perhaps there is a time lag between me and the authors, and the missed data seems to contain data less than a day. If the author got the data at Chinese time, there may be a data difference. The data differences may lead to a difference in the final result.

\subsubsection{RL Stochasticity}
The different initial settings and training epochs may also cause the difference. Although the paper provided the parameters of the model, the author did not give more details about the training, such as the seed and replay buffer history. Also, the epochs for training are not announced in the paper. In the paper, the author said they used early stopping, but did not talk about the loss details, stop epochs and other parameters. 

Based on my experiments, the model is very computationally resource-consumption due to the DQN network and renewal methods. We trained on the laptop with RTX 2070, which needs more than 20 minutes for one epoch, so it is less possible for me to entirely train the group and fine-tune the model during a week. So these may also cause the differences.

\subsubsection{Differences for Baselines}
\label{baseline diff}
There are some differences between the baseline methods. For M2, the setting of the range about the price may cause a difference that some people may consider the tick and others consider the mid price or oracle price. These will lead to the difference between the gas fee and the trading fee (so you can see the PnL of M2 performs closer to the original result than others). This situation is also the same for M3.

For M4, we think there is a mistake by the author that DP is more suitable for a continuous model, but not a discrete environment. Based on the assumption of \cite{cartea2024decentralized}, the price should be a GBM in continuous time, which may not be very suit to the assumptions in this paper, so the result of M4 may be not very reliable.

\subsubsection{Unmentioned Protocols}
There are some unclear instructions in the paper, and some of the missing instructions make me confused during the replication. Firstly, although there is an assumption about the "flat" gas fee in the paper, there are no clear instructions about which gas fee was used in the experiments and it is a very important parameter for the replication. 

Another confusing instruction is that the paper talked less about the details of the DDQN with and without "hedging". And it is confusing that none of the results of PnL shown in Table 5 (0.223 for period 1, $l_0 =250$; 0.183 for period 2, $l_0 =250$) in the original paper are close to the results in Table 6 (0.373 for period 1, $l_0 =250$; 0.307 for period 2, $l_0 =250$). 

These actually influenced some of the results. We had emailed the authors, but had not waited for the reply till the date we submitted. We will try other ways to fix or connect to the authors for more information, if possible, for further work.

\section{Extension: Mamba DDQN in Uniswap V3}
\label{Extension}
%What you changed or added: your extension idea (for example, a new reward function, volatility-sensitive policy, or risk constraint).
\subsection{Motivation}
Based on the experience during the replication and comments in the Section \ref{comment}, We find some limitations with the original work. First, the original paper directly used 28 features along with OHLCV-based features to build a 32-dimensional input vector. However, the features seem neither screened nor ranked, and the data are not standardized. These may influence the quality of the model and training.

Second, Dueling DDQN uses a simple 2-layer MLP (64-64) as the feature extractor. However, DeFi data shows the strong temporal dependencies, regime shifts, and non-local interactions across features. MLPs may fail to capture these because they lack sequence modeling capability. However, State-Space-Model (SSM), such as Mamba \cite{gu2024mamba} is able to capture long-range dependencies efficiently, maintain stability during training and is more robust to non-stationary and noisy data. So it is a potential idea to use Mamba instead of MLP.

Another point is that some baseline methods used in previous work may not be well-suited and well-reflect the performance of the model. We had discussed this in the Section \ref{baseline diff}. So it is necessary to provide some clearer baselines to be compared with.

Based on these observations and analysis, we propose a new extended structure named Mamba DDQN that uses Mamba instead of an MLP layer with a new reward function. We will start with data cleaning that optimizes the number of features used in the model. Then we will normalize the data when dividing the dataset. For the baselines, we use two new baselines: Buy-and-Hold and Daily Rebalancing. These baselines are clearer and simpler, which may be better baselines to compare with.

\subsection{Data Processing}
\begin{figure}[htbp]
    \centering
    
    \begin{subfigure}{0.42\linewidth}
        \centering
        \includegraphics[width=\linewidth]{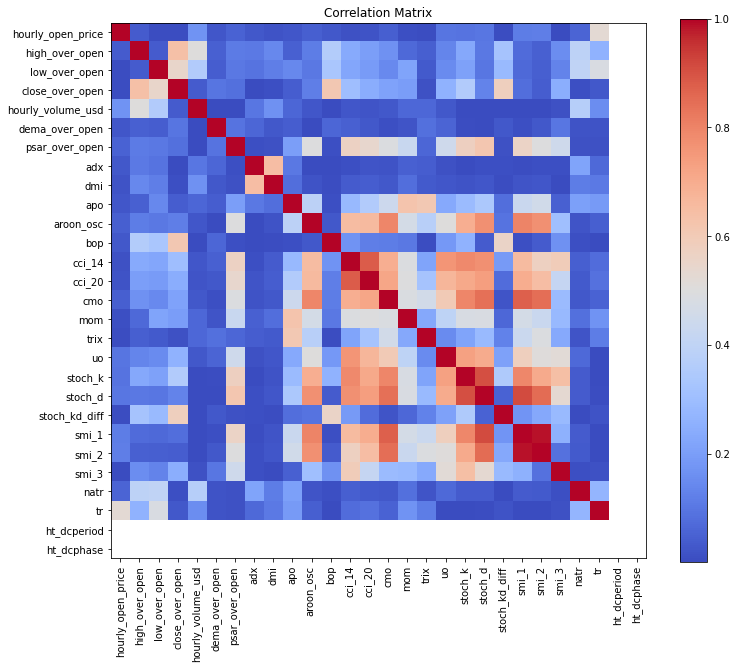}
        \caption{Correlation between Features}
        \label{fig:correlation}
    \end{subfigure}
    \hfill
    \begin{subfigure}{0.48\linewidth}
        \centering
        \includegraphics[width=\linewidth]{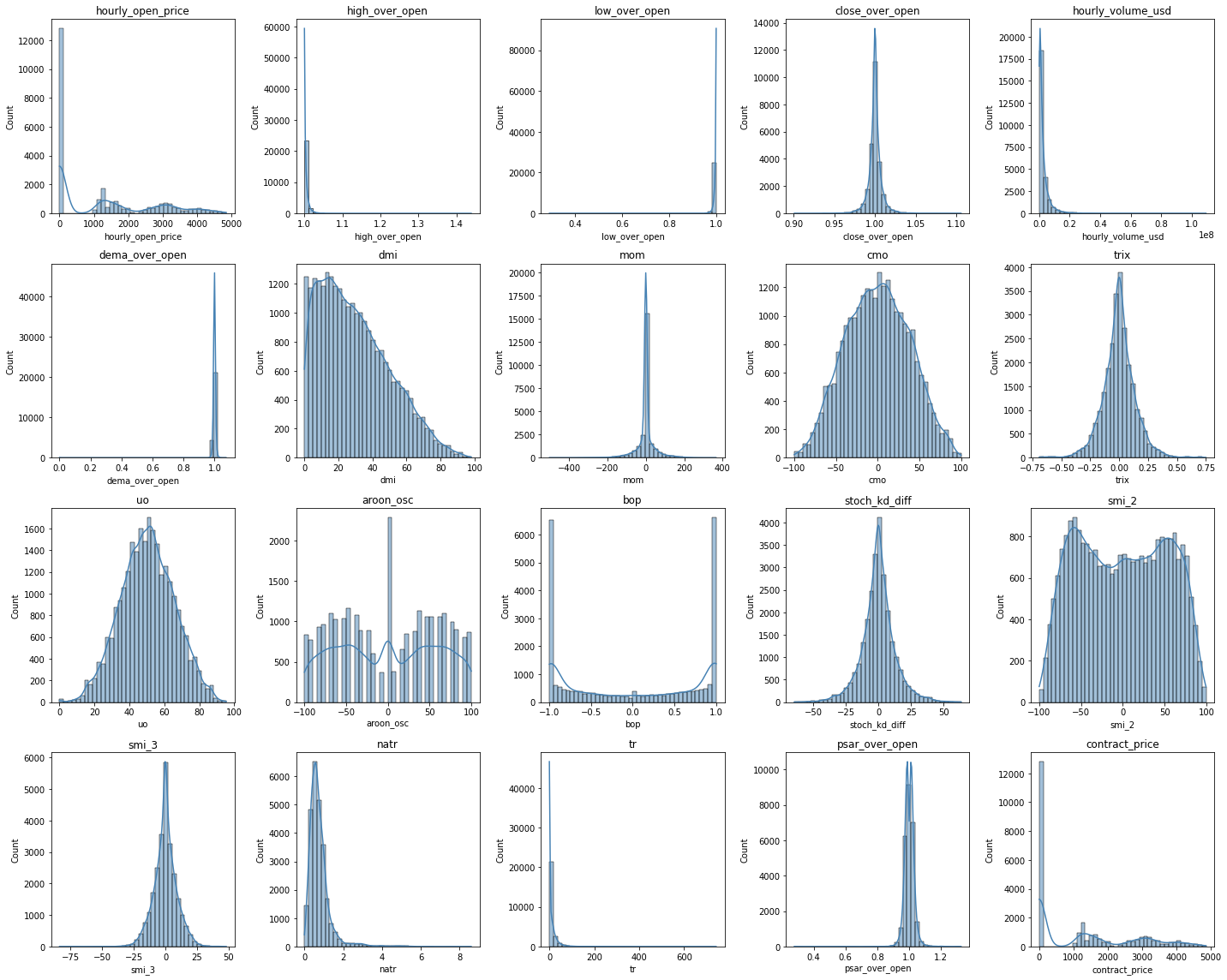}
        \caption{Distributions of Features}
        \label{fig:distribution}
    \end{subfigure}

    \caption{Correlation and distribution of features}
    \label{fig:twofigs}
\end{figure}

For the first step, we first do a correlation test for every feature in the dataset. The result of the correlation is shown in Figure \ref{fig:correlation}. Set the threshold as 0.8, there are some features that show high correlations with other features: $'cmo'$, $'smi\_1'$, $'stoch\_d'$, $'aroon\_osc'$, $'cci\_14'$, $'stoch\_k'$.

Then we select the features by some traditional machine learning methods, which are Lasso, ElasticNet, Random Forest, and XGBoost. Here is the features that models suggest to save:

\textbf{Lasso}: $['hourly\_open\_price', 'high\_over\_open', 'low\_over\_open', 'close\_over\_open', 'hourly\_volume\_usd', \\
'dema\_over\_open', 'dmi', 'aroon\_osc', 'bop', 'cci\_14', 'cci\_20', 'cmo', 'mom', 'trix', 'uo', 'stoch\_kd\_diff', \\
'smi\_2', 'smi\_3', 'natr', 'tr']$

\textbf{ElasticNet}:$['hourly\_open\_price', 'high\_over\_open', 'low\_over\_open', 'close\_over\_open', 'hourly\_volume\_usd', \\
'dema\_over\_open', 'dmi', 'aroon\_osc', 'bop', 'cci\_14', 'cci\_20', 'cmo', 'mom', 'trix', 'uo', 'stoch\_kd\_diff', 'smi\_2', \\'smi\_3', 'natr', 'tr']$

\textbf{Random Forest}: $['low\_over\_open', 'close\_over\_open', 'hourly\_volume\_usd', 'dema\_over\_open', 'natr' \\'psar\_over\_open', 'dmi', 'bop', 'trix', 'smi\_3']$

\textbf{XGBoost}: $['hourly\_open\_price', 'low\_over\_open', 'close\_over\_open', 'hourly\_volume\_usd', 'psar\_over\_open', \\'dmi', 'bop', 'cci\_14', 'cci\_20', 'cmo']$

we give the model that only one or none of the models are monitored and with high correlations with other features. After the pruning, the final given-up features are: $"adx"$, $"apo"$, $"ht\_dcperiod"$, $"ht\_dcphase"$, $"cci\_20"$, $"smi\_1"$, $"stoch\_k"$, $"stoch\_d"$. In this way, there are 20 features for training now.

After the pruning of the data, we will take normalization for the last dataset. Just as Figure \ref{fig:distribution}, the distributions of many features are not Gaussian, and many of them have a heavy-tail or extreme values. For features about volume and price, we take a log-transform. And all the features will take a Z-score normalization:
$$
z_t = \frac{x_t - \mu_{train}}{\sigma_{train}}
$$

\subsection{Methodology}
The framework of the Mamba DDQN is shown in Figure \ref{Framework of DDQN}. Mamba \cite{gu2024mamba} is a kind of SSMs with series states and is suitable for financial noisy modeling. This subsetion will mainly introduce the difference between Mamba DDQN and Dueling DDQN.
\begin{figure}
    \centering
    \includegraphics[width=\linewidth]{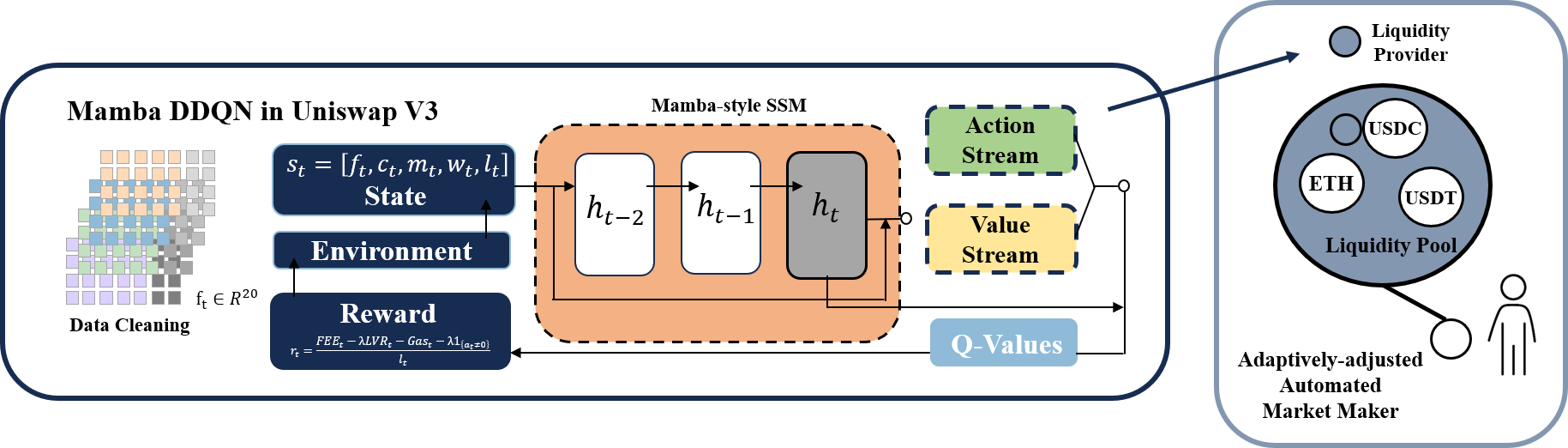}
    \caption{Framework of Mamba DDQN}
    \label{Framework of Mamba DDQN}
\end{figure}

\subsubsection{State and Replay Buffer}
Recall the state in Dueling DDQN: $s_t = [\textbf{f}_t, c_t, m_t, w_t, l_t]$. Then in the Mamba-DDQN, there is a fixed-length historical series for $s_t$, which is:
$$
S_t = [s_{t-31}, \ldots, s_t] \in \mathbf{R}^{32\times D}
$$
Based on this type of state, the model can store the historical series for reference. It is better for time series data. And a 32-dimensional state replay buffer will be used instead of the single-dimensional buffer in DDQN. 
\begin{lstlisting}[basicstyle=\ttfamily\small]
state_seq   : (T, state_dim)
next_state_seq : (T, state_dim)
state_buffer = deque(maxlen=HISTORY_LEN)
np.stack(list(state_buffer))
\end{lstlisting}

\subsubsection{Reward}
In the new model, we add a risk penalty based on the original model. In the original model, the model will always do nothing because there are fewer penalties for the action for a long time. So we introduced a penalty parameter $\lambda$ and the new reward function is:
$$
r_t = \frac{Fee_t - \lambda\cdot LVR_t - Gas_t - \lambda \cdot\textbf{1}_{\{a_t \neq a_{t-1}\}}}{l_0}
$$

In this way, the agent will try to take action more frequently than it will receive a penalty if it has not taken action for a long time.

\subsubsection{Mamba Layer}
In the Mamba-DDQN, we add a simple SSM with Mamba-style, which is:
\begin{align}
    h_{t+1} &= GELU(h_{t-1}A+x_tB)\\
   & = GELU(GELU(h_{t-2}A+x_{t-1}B)A + x_tB)\\
   & = F(x_{t-T+1}, \ldots, x_t)
\end{align}

Where with a long-term memory:
$$
h_t = \sum^T_{k=0}x_{t-k}BA^k
$$

In the code, it implemented in this way:
\begin{lstlisting}[basicstyle=\ttfamily\small]
h_{t+1} = GELU(h_t A + x_t B)
h = torch.matmul(h, self.A) + torch.matmul(x_t, self.B)
h = gelu(h)
\end{lstlisting}

And finally the Q-funtion:
$$
V(S_t), A(S_t,a) = SSM + Linear(S_t)
$$

\subsection{Result}
%A comparison between your extended model and the original results (how did 
With the same parameters of the original model, we trained the model, and the result is shown in the Table \ref{MDQN}, where M-DDQN is the Mamba-DDQN.

\begin{table}[htbp]
\centering
\caption{Relative PnL of M1 and Mamba-DDQN w/ and w/o Hedging }
\begin{tabular}{l l r r r| r r r}
\hline
 & & \multicolumn{3}{c}{ETH/USDC-0.3\%} & \multicolumn{3}{c}{ETH/USDT-0.3\%} \\
 & & M-DDQN & M1 & M1$^\dagger$ & M-DDQN & M1 & M1$^\dagger$ \\
\hline
\multirow{4}{*}{$l_0 = 250$}
  & period 1 & 0.025 & \textbf{0.223} & -0.259 & 0.022 & \textbf{0.186} & -0.259 \\
  & period 2 & 0.073 & \textbf{0.183} &  0.034 & 0.061 & \textbf{0.151} &  0.079 \\
  & period 3 & 0.075 & \textbf{0.130} & -0.131 & 0.054 & \textbf{0.139} & -0.141 \\
  & period 4 & 0.104 & \textbf{0.110} &  0.062 & 0.038 & \textbf{0.074} &  0.029 \\
\hline
\multirow{4}{*}{$l_0 = 500$}
  & period 1 & 0.197 & \textbf{0.305} & -0.291 & 0.228 & \textbf{0.275} & -0.270 \\
  & period 2 & 0.092 & \textbf{0.273} &  0.043 & 0.153 & \textbf{0.267} &  0.022 \\
  & period 3 & \textbf{0.386} & 0.302 & -0.178 & \textbf{0.312} & 0.195 & -0.125 \\
  & period 4 & \textbf{0.167} & 0.159 &  0.023 & \textbf{0.190} & 0.143 &  0.080 \\
\hline
\multirow{4}{*}{$l_0 = 1000$}
  & period 1 & \textbf{0.946} & 0.582 & -0.283 & \textbf{0.948} & 0.525 & -0.278 \\
  & period 2 & 0.364 & \textbf{0.388} &  0.061 & 0.370 & \textbf{0.383} &  0.050 \\
  & period 3 & 0.184 & \textbf{0.400} & -0.119 & 0.181 & \textbf{0.354} & -0.105 \\
  & period 4 & 0.068 & \textbf{0.222} &  0.079 & 0.083 & \textbf{0.189} &  0.071 \\
\hline
\end{tabular}

\vspace{0.3em}

{\footnotesize
M1$^\dagger$ means Dueling DDQN is trained with the reward function
defined as PnL when hedging is not used}
\label{MDQN}
\end{table}

Based on the result, we can find that M-DDQN performs better than the Dueling-DDQN without hedging in all the sets and periods, but only better than Dueling-DDQN with hedging in a few of the sets and periods. When $l_0$ =250, M-DDQN seems very careful and acts less than others (as shown in trading fees). As the value of $l_0$ increases, the PnL of M-DDQN performs better and sometimes gains a much higher PnL than others when $l_0=1000$. M-DDQN expresses better robustness and cross-period consistency and is more stable in returns and improved risk control.

\subsection{New Baselines}
To better evaluate the performance of the models, two new baselines are used in the extended project. The first baseline is the buy-and-hold strategy, which is to keep holding the price of ETH during the period. Another one is the daily rebalancing, which is to rebalance the position every day. The results of adding new baselines are shown in Tables \ref{tab:performance11} and \ref{tab:performance22} for the pools of ETH/USDC-$0.3\%$ and ETH/USDT-$0.3\%$, respectively. We can see that the Buy-and-Hold strategy is highly influenced by the market risk, so the PnL is not very stable. The daily rebalancing strategy always pays a high amount of trading fees and gas fees because it changes the position everyday.

\begin{table*}[ht]
\centering
\small
\begin{tabular}{llcccc|cc}
\toprule
& \textbf{ETH/USDC-0.3\%} & \multicolumn{4}{c}{\textbf{Original Result}} & \multicolumn{2}{c}{\textbf{New Baseline}} \\
\cmidrule(lr){3-6} \cmidrule(lr){7-8}
& & M1 & M2 & M3 & M4 & Buy-and-Hold & Daily Rebalance \\
\midrule

\textbf{Period 1} & relative trading fee & 0.691 & 0.630 & \textbf{0.774} & 0.088 &  / & 0.544 \\
& relative gas fee     & \textbf{0.113} & 0.207 & 0.293 & 0.120 & / & 0.200 \\
& relative LVR         & 0.205 & 0.205 & 0.248 & \textbf{0.030} & / & 0.460 \\
& relative PnL         & 0.373 & 0.218 & 0.232 & -0.210 & \textbf{0.508} & -0.022\\

\midrule
\textbf{Period 2} & relative trading fee & 0.607 & 0.453 & \textbf{0.611} & -0.052 & / & 0.372 \\
& relative gas fee     & \textbf{0.118} & 0.153 & 0.253 & 0.120 & / & 0.205 \\
& relative LVR         & 0.182 & 0.132 & 0.168 & \textbf{0.015} & / & 0.221 \\
& relative PnL         & \textbf{0.307} & 0.168 & 0.190 & -0.083 & -0.188 & -0.163 \\

\midrule
\textbf{Period 3} & relative trading fee & 0.541 & 0.569 & \textbf{0.700} & 0.089 & / & 0.326 \\
& relative gas fee     & \textbf{0.104} & 0.167 & 0.307 & 0.124 & / & 0.200 \\
& relative LVR         & 0.220 & 0.257 & 0.174 & \textbf{0.045} & / & 0.240 \\
&relative PnL         & \textbf{0.217} & 0.145 & 0.096 & -0.079 & 0.162 & -0.193 \\

\midrule
\textbf{Period 4} & relative trading fee & 0.370 & 0.318 & \textbf{0.365} & 0.031 & / &0.208 \\
& relative gas fee     & \textbf{0.064} & 0.100 & 0.253 & 0.120 & / & 0.190 \\
& relative LVR         & 0.121 & 0.108 & 0.107 & \textbf{0.014} & / & 0.145 \\
& relative PnL         & \textbf{0.185} & 0.109 & 0.004 & -0.104& -0.158 & -0.166 \\

\bottomrule
\end{tabular}
\caption{Performance of M1--M4 across Four Periods for ETH/USDC-0.3\% With New Baselines}
\label{tab:performance11}
\end{table*}

\begin{table*}[]
\centering
\small
\begin{tabular}{llcccc|cc}
\toprule
& \textbf{ETH/USDT-0.3\%} & \multicolumn{4}{c}{\textbf{Original Result}} & \multicolumn{2}{c}{\textbf{New Baseline}} \\
\cmidrule(lr){3-6} \cmidrule(lr){7-8}
& & M1 & M2 & M3 & M4 & Buy-and-Hold & Daily Rebalance\\
\midrule

\textbf{Period 1} & relative trading fee & 0.645 & 0.614 & \textbf{0.921} & 0.402 & / & 0.464 \\
& relative gas fee     & \textbf{0.096} & 0.207 & 0.293 & 0.133 & / & 0.200 \\
& relative LVR         & 0.199 & 0.211 & 0.316 & \textbf{0.139} & / & 0.390 \\
& relative PnL         & \textbf{0.310} & 0.196 & 0.131 & 0.130 & -0.337 & -0.348 \\

\midrule
\textbf{Period 2}  & relative trading fee & 0.503 & 0.459 & \textbf{0.587} & 0.452 & / & 0.416 \\
& relative gas fee     & \textbf{0.089} & 0.147 & 0.253 & 0.140 & / & 0.205 \\
& relative LVR         & 0.161 & 0.159 & 0.186 & \textbf{0.125} & / & 0.255 \\
& relative PnL         & \textbf{0.252} & 0.154 & 0.178 & 0.187 & 0.223 & 0.054\\

\midrule
\textbf{Period 3} & relative trading fee & 0.686 & 0.707 & \textbf{0.774} & 0.295 & / & 0.252 \\
& relative gas fee     & \textbf{0.119} & 0.060 & 0.207 & 0.094 & / & 0.191 \\
& relative LVR         & 0.335 & 0.209 & 0.314 & \textbf{0.103} & / & 0.183 \\
& relative PnL         & \textbf{0.232} & 0.122 & 0.109 & 0.097 & -0.140 & -0.335\\

\midrule
\textbf{Period 4} & relative trading fee & 0.463 & 0.348 & \textbf{0.373} & 0.149 & / & 0.226 \\
& relative gas fee     & \textbf{0.066} & 0.107 & 0.293 & 0.120 & / & 0.205 \\
& relative LVR         & 0.271 & 0.159 & 0.115 & \textbf{0.059} & / & 0.131 \\
& relative PnL         & 0.124 & 0.082 & -0.035 & -0.030 & \textbf{0.186} & -0.028 \\

\bottomrule
\end{tabular}
\caption{Performance of M1--M4 across Four Periods for ETH/USDT-0.3\% With New Baselines}
\label{tab:performance22}
\end{table*}

\section{Discussion}
\label{Discussion}
%Your reflections on why this extension matters and how it could fit into future research in the QGI lab.
The extension works still have many potential improvements for future development. For example, we can use longer training horizons that may further improve generalization. We can use more efficient training algorithms \cite{zhang2024quantformer} or faster sequence models could address the current computational bottleneck. One week seems not enough for me to train all the models very efficiently and fine-tune. Adding more refined reward shaping and penalty design may be a good idea to improve the performance of M-DDQN.

%For future work, aligning with the QGI Lab, we think it is a good fit. Within the Geometry of Financial Markets theme, this project links to how high-frequency, noisy and non-stationary market DeFi data can be structured and modeled, such that Mamba-style models can effectively capture. For the Geometric Foundations of Decision-Making direction, this report provides an example of how to use deep reinforcement learning agents to make stable and robust decisions under uncertainty and high-dimensional information with the geometric structure of decision spaces. The QGI Lab is such a very young, energetic group that has great potential for future research. we will be honored if we can get into the lab and show my ability.

\bibliographystyle{plain}
\bibliography{main}

\end{document}